# A Predictive System for detection of Bankruptcy using Machine Learning techniques

Kalyan Nagaraj[1*] and Amulyashree Sridhar[2]

[1*]PES Institute of Technology
kalyan1991n@gmail.com

[2]PES Institute of Technology
0908amulyashree@gmail.com

## Abstract

Bankruptcy is a legal procedure that claims a person or organization as a debtor. It is essential to ascertain the risk of bankruptcy at initial stages to prevent financial losses. In this perspective, different soft computing techniques can be employed to ascertain bankruptcy. This study proposes a bankruptcy prediction system to categorize the companies based on extent of risk. The prediction system acts as a decision support tool for detection of bankruptcy

# 1. Introduction

Bankruptcy is a situation in which a firm is incapable to resolve its monetary obligations leading to legal threat. The financial assets of companies are sold out to clear the debt which results in huge financial losses to the investors. Bankruptcy results in decreased liquidity of capital and minimized financial improvement. It is reported by World Bank data that Indian government resolves the insolvency in an average of 4.3 years [1]. There is a need to design effective strategies for prediction of bankruptcy at an earlier stage to avoid financial crisis. Bankruptcy can be predicted using mathematical techniques, hypothetical models as well as soft computing techniques [2]. Mathematical techniques are primary methods used for estimation of bankruptcy based on financial ratios. These methods are based on single or multi variable models. Hypothetical models are developed to support the theoretical principles. These models are statistically very complex based on their assumptions. Hence soft computing techniques are extensively used for developing predictive models in finance. Some of the popular soft computing techniques include Bayesian networks, logistic regression, decision tress, support vector machines and neural networks.

In this study, different soft computing techniques are employed to predict bankruptcy. Further based on the performance of the classifiers, the best model is chosen for development of a decision support system in R programming language. The support system can be utilized by stock holders and investors to predict the performance of a company based on the nature of risk associated.

# 2. Background

Several studies have been conducted in the recent past reflecting the importance of soft computing techniques. The studies and the technologies implemented are briefly discussed below.

## 2.1. Machine Learning

Machine learning techniques are employed to explore the hidden patterns in data by developing models. It is broadly referred as knowledge discovery in database (KDD). Different learning algorithms are implemented to extract patterns from data. These algorithms can either be supervised or unsupervised. Supervised learning is applied when the output of a function is previously known. Unsupervised learning is applied when the target function is unknown. The general layout for machine learning process is described below:

*Data collection*: The data related to domain of concern is extracted from public platforms and data warehouses. The data will be raw and unstructured format. Hence pre-processing measures must be adopted

*Data pre-processing*: The initial dataset is subjected for pre-processing. Pre-processing is performed to remove the outliers and redundant data. The missing values are replaced by normalization and transformation

*Development of models*: The pre-processed data is subjected to different machine learning algorithms for development of models. The models are constructed based on classification, clustering, pattern recognition and association rules

*Knowledge Extraction:* The models are evaluated to represent the knowledge captured. This knowledge attained can be used for better decision making process [3].

## 2.2 Classification algorithms

Several classification algorithms are implemented in recent past for financial applications. They are discussed briefly below:

**Logistic Regression**: It is a classifier that predicts the outcome based probabilities of logistic function. It estimates the relationship between different independent variables and the dependent outcome variable based on probabilistic value. It may be either binary or multinomial classifier. The logistic function is denoted as:

$$F(x) = \frac{1}{1+e^{-(\beta_0+\beta_1 x)}}$$

$\beta_0$ and $\beta_1$ are coefficients for input variable x. The value of F(x) ranges from zero to one. The logistic regression model generated is also called as generalized linear model [4].

**Naïve Bayes classifier:** It is a probabilistic classifier based on the assumptions of Bayes theorem [5]. It is based on independent dependency among all the features in the dataset. Each feature contributes independently to the total probability in model. The classifier is used for supervised learning. The Bayesian probabilistic model is defined as:

$$p(C_k \mid x) = \frac{p(C_k) p(x \mid C_k)}{p(x)}$$

$p(C_k|x)$ = posterior probability

$p(C_k)$ = prior probability

$p(x)$ = probability of estimate

$p(x|C_k)$ = likelihood of occurrence of x

**Random Forest**: They are classifier which construct decision trees for building the model and outputs the mode value of individual trees as result of prediction. The algorithm was developed by Breiman [6]. Classification is performed by selecting a new input vector from training set. The vector is placed at the bottom of each of the trees in the forest. The proximity is computed for the tree. If the tree branches are at the same level, then proximity is incremented by one. The proximity evaluated is standardized as a function of the number of trees generated. Random forest algorithms compute the important features in a dataset based on the out of bag error estimate. The algorithm also reduces the rate of overfitting observed in decision tree models.

**Neural networks**: They are learning algorithms inspired from the neurons in human brain. The network comprises of interconnected neurons as a function of input data [7]. Based on the synapse received from input data, weights are generated to compute the output function. The networks can either be feed-forward or feed-back in nature depending upon the directed path of the output function. The error in input function is minimized by subjecting the network for back-propagation which optimizes the error rate. The network may also be computed from several layers of input called as multilayer perceptron. Neural networks have immense applications in pattern recognition, speech recognition and financial modeling.

**Support vector machine**: They are supervised learning algorithms based on non-probabilistic classification of dataset into categories in high dimensional space. The algorithm was proposed by Vapnik [8]. The training dataset is assumed as a p-dimensional vector which is to be classified using (p-1) dimensional hyperplane. The largest separation achieved between data points is considered optimal. Hyperplane function is represented as:

$$f(x, <w, b>) = sign(w.x + b)$$

w = normalized vector to the hyperplane
x = p-dimensional input vector
b = bias value
The marginal separator is defined as $2|k|/||w||$. 'k' represents the number of support vectors generated by the model. The data instances are classified based on the below criteria
If $(w \cdot x + b) = k$, indicates all the positive instances.
If $(w \cdot x + b) = -k$, indicates the set of negative instances.
If $(w \cdot x + b) = 0$, indicates the set of neutral instances.

## 3. Related Work

Detection of bankruptcy is a typical classification problem in machine learning application. Development of mathematical and statistical models for bankruptcy prediction was initiated by Beaver in the year 1960 [9]. The study focused on the univariate analysis of different financial factors to detect bankruptcy. An important development in this arena was recognized by Altman who developed a multivariate Z-score model of five variables [10]. Z-score model is considered as a standard model for estimating the probability of default in bankruptcy. Logistic regression was also instigated to evaluate bankruptcy [11]. These techniques are considered as standard estimates for prediction of financial distress. But these models pose statistical restrictions leading to their limitations. To overcome these limitations probit [12] and logit models [13] were implemented for financial applications. In later years, neural networks were implemented for estimating the distress in financial organizations [14, 15, and 16]. Neural networks are often subjected to overfitting leading to false predictions. Decision trees were also applied for predicting financial distress [17, 18]. Support vector machines have also been used employed in predicting bankruptcy for financial companies [19, 20]. In recent years, several hybrid models have been adopted to improve the performance of individual classifiers for detection of bankruptcy [21, 22].

## 4. Methodology

### 4.1. Collection of Bankruptcy dataset
The qualitative bankruptcy dataset was retrieved from UCI Machine Learning Repository [23]. The dataset comprised of 250 instances based on 6 attributes. The output had two classes of nominal type describing the instance as 'Bankrupt' (107 cases) or 'Non-bankrupt' (143 cases).

### 4.2. Feature Selection

It is important to remove the redundant attributes from the dataset. Hence correlation based feature selection technique was employed. The feature based algorithm selects the significant attributes based on the class value. If the attribute is having high correlation with the class variable and minimal correlation with other attributes of the dataset it is presumed to be a good attribute. If the attribute have high correlation with the attributes then they are discarded from the study.

### 4.3. Implementing machine learning algorithms

The features selected after correlational analysis are subjected to data partitioning followed by application of different machine learning algorithms. The dataset is split into training ($2/3^{rd}$ of the dataset) and test dataset ($1/3^{rd}$ of the dataset) respectively. In the training phase different classifiers are applied to build an optimal model. The model is validated using the test set in the testing phase. Once the dataset is segregated, different learning algorithms was employed on the training dataset. The algorithms include logistic regression, Bayesian classifier, random forest, neural network and support vector machines. Models generated from each of the classifier were assessed for their performance using the test dataset. A ten-fold cross validation strategy was adopted to test the accuracy. In this procedure, the test dataset is partitioned into ten subsamples and each subsample is used to test the performance of the model generated from training dataset. This step is performed to minimize the probability of overfitting. The accuracy of each algorithm was estimated from the cross validated outcomes.

### 4.4. Developing a predictive decision support system

From the previous step, the classifier with highest prediction accuracy is selected for developing a decision support system to predict the nature of bankruptcy. The prediction system was implemented in RStudio interface, a statistical programming toolkit. Different libraries were invoked for development of the predictive system including 'gWidgets 'and 'RGtk2'. The predictive tool develops a model for evaluating the outcome bankruptcy class for user input data. Predicted class is compared with the actual class value from the dataset to compute the percentage of error prediction from the system. The support system estimates the probability of bankruptcy among customers. It can be used as an initial screening tool to strengthen the default estimate of a customer based on his practises.

The methodology of this study is illustrated in Figure 1.

**Figure 1**: The flowchart for developing a decision support system to predict bankruptcy

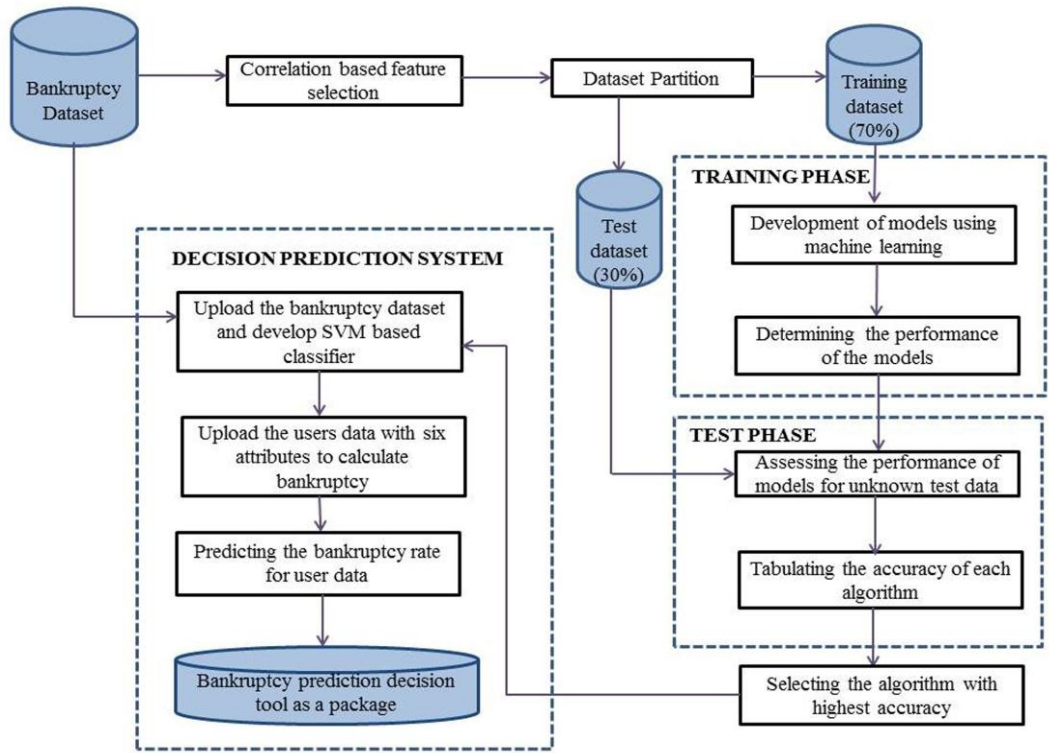

## 5. Results and Discussion

### 5.1. Description of Qualitative Bankruptcy dataset

The bankruptcy dataset utilized for this study is available at UCI Machine Learning Repository. The dataset comprising of six different features is described in Table 1. The distribution of class outcome is shown in Figure-2.

Table 1: Qualitative Bankruptcy Dataset. (Here P=Positive, A=Average, N=Negative, NB=Non-Bankruptcy and B=Bankruptcy)

| Sl. No | Attribute Name | Description of attribute |
|--------|----------------|--------------------------|
| 01. | IR (Industrial Risk) | Nominal {P, A, N} |
| 02. | MR (Management Risk) | Nominal {P, A, N} |
| 03. | FF (Financial Flexibility) | Nominal {P, A, N} |
| 04. | CR (Credibility) | Nominal {P, A, N} |
| 05. | CO (Competitiveness) | Nominal {P, A, N} |
| 06. | OR (Operating Risk) | Nominal {P, A, N} |
| 07. | Class | Nominal {NB, B} |

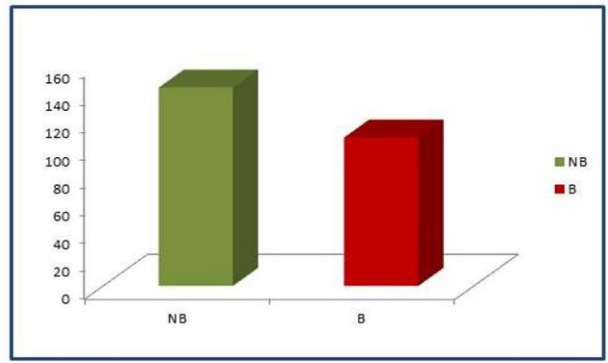

Figure 2: The distribution of class output representing Non Bankruptcy and Bankruptcy

### 5.2. Correlation based Attribute selection

The bankruptcy dataset was subjected for feature selection to extract the relevant attributes. The nominal values in bankruptcy dataset were converted to numeric values for performing feature selection. The values of each of the descriptors were scaled as P=1, A=0.5 and N=0 representing the range for positive, average and negative values. The procedure was repeated for all the six attributes in dataset. Pearson correlation filter was applied for the numerical dataset to remove the redundant features with a threshold value of 0.7. The analysis revealed that all the six attributes were highly correlated with the outcome variable. In order to confirm the results from correlation, another feature selection method was applied for the dataset. Information gain ranking filter method was applied to test the importance of features. The algorithm discovered similar results as that of correlation. Hence all the six attributes from the dataset were considered for the study. The correlational plot for the features is shown in Figure 3.

Figure 3: The correlational plot illustrating the importance of each feature

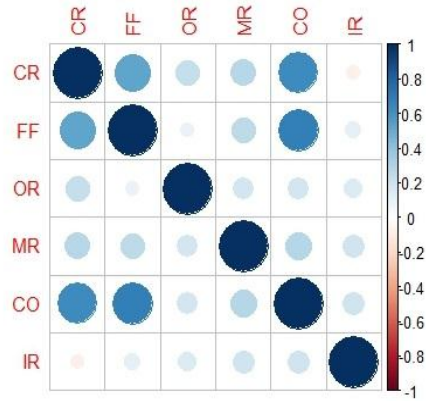

**5.3. Machine learning algorithms**

The features extracted from previous step were subjected for different machine learning algorithms in R. The algorithms were initially applied for the training set to develop predictive models. These models were further evaluated using the test set. Performance of each model was adjudged using different statistical parameters like confusion matrix and receiver operating characteristics (ROC) curve. Confusion matrix is a contingency table that represents the performance of machine learning algorithms [24]. It represents the relationship between actual class outcome and predicted class outcome based on the following four estimates:

a) True positive (TP): The actual negative class outcome is predicted as negative class from the model

b) False positive (FP): The actual negative class outcome is predicted as a positive class outcome. It leads to Type-1 error

c) False negative (FN): The actual positive class outcome is predicted as negative class from the model. It leads to Type-2 error

d) True negative (TN): The actual class outcome excluded is also predicted to be excluded from the model.

Based on these four parameters the performance of algorithms can be adjudged by the following ratios.

$$Accuracy(\%) = \frac{TP+TN}{TP+FP+TN+FN}$$

$$TPR(\%) = \frac{TP}{TP+FN}$$

$$FPR(\%) = \frac{FP}{FP+TN}$$

$$precision(\%) = \frac{TP}{TP+FP}$$

ROC curve is a plot of false positive rate (X-axis) versus true positive rate (Y-axis). It is represents the accuracy of a classifier [25].

The accuracy for all the models was computed and represented in Table 2. The ROC plot of SVM classifier is represented in Figure 4.

Table 2: The accuracy of bankruptcy prediction of machine learning algorithms

| Sl. No | Algorithm | Library used in R | Accuracy of prediction (%) | True positive rate | False positive rate | Precision |
|---|---|---|---|---|---|---|
| 01. | Logistic regression | glmnet | 97.2 | 0.972 | 0.028 | 0.97 |
| 02. | Rotation forest | randomForest | 97.4 | 0.974 | 0.026 | 0.97 |
| 03. | Naïve Bayes | e1071 | 98.3 | 0.983 | 0.017 | 0.98 |
| 04. | Neural network | neuralnet | 98.6 | 0.986 | 0.014 | 0.98 |
| 05. | RBF-based Support vector machine | e1071 | 99.6 | 0.996 | 0.004 | 0.99 |

Figure 4: The ROC curve representing the accuracy of SVM classifier

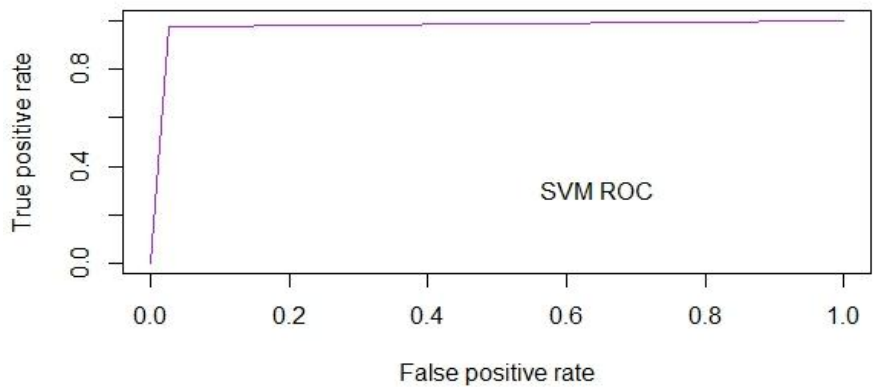

## 5.4. SVM based decision supportive system in R

Based on the accuracy in previous step, it was seen that support vector based classifier outperformed other machine learning techniques. The classifier was implemented using radial basis function (RBF) kernel. It is also referred as Gaussian RBF kernel. The kernel representation creates a decision boundary for the non-linear attributes in high dimensional space. The attributes are converted to linear form by mapping using this kernel function. An optimal hyperplane is constructed in feature space by considering the inner product of the kernel. Hyperplane is considered as optimal if it creates a widest gap from the input attributes to the target class. Furthermore, to achieve optimization C and gamma parameters are used. C is used to minimize the misclassification in training dataset. If the value of C is smaller it is soft margin creating a wider hyperplane, whereas the value of C being larger leads to overfitting called as hard margin. Hence the value of C must be selected by balancing between the soft and hard margin. Gamma is used for non-linear classifiers for constructing the hyperplane. It is used to control the shape of the classes to be separated. If the value of gamma is small it results in high variance and minimal bias producing a pointed thrust. While a bigger gamma value leads to minimal variance and maximum bias producing a broader and soft thrust. The values of C and gamma were optimized and selected for classification. The classified instances from RBF kernel is observed in Figure 5.

Figure 5: RBF classifier based classification for bankruptcy dataset as either NB or B

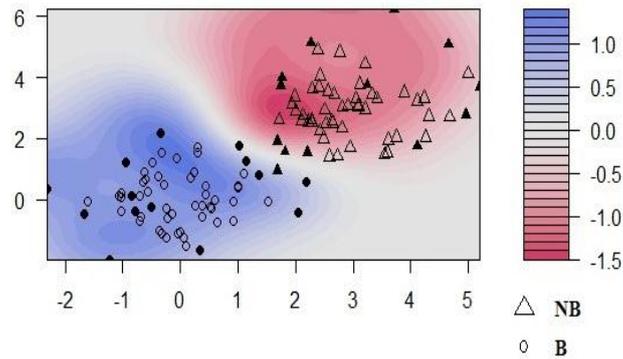

Based on the RBF classifier the prediction system was constructed in R. The bankruptcy dataset is initially loaded into the predictive system as a .csv file. The home page of predictive tool loaded with bankruptcy dataset is shown in Figure 6.

Figure 6: SVM based predictive tool with the bankruptcy dataset

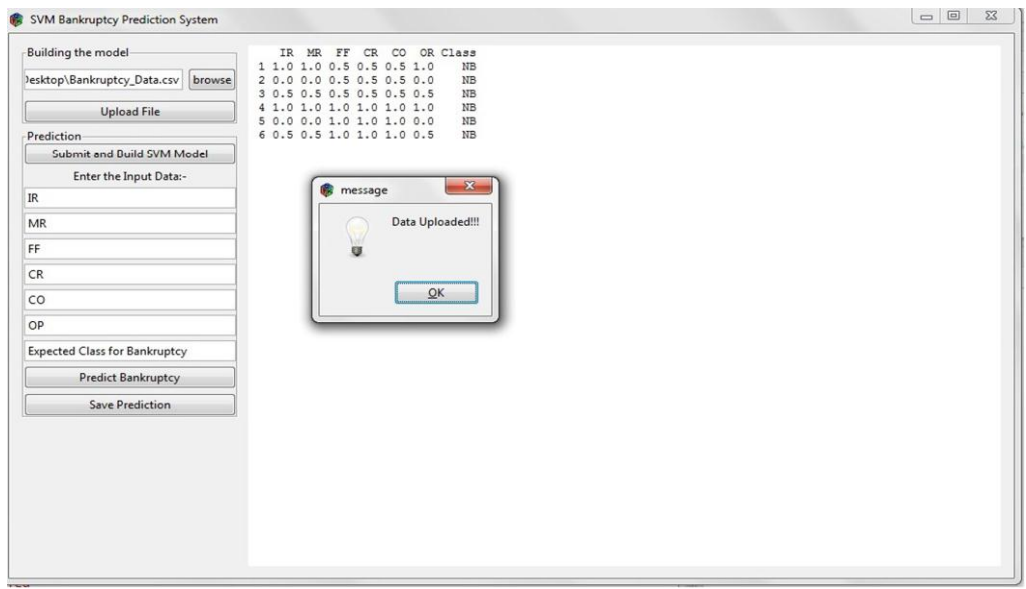

The system fetches the dataset and stores as a dataframe. Dataframe is a vector list used to store the data as a table in R. RBF-kernel SVM model is developed for the bankruptcy dataset. It is displayed in Figure 7.

Figure 7: Radial based SVM model developed for bankruptcy dataset

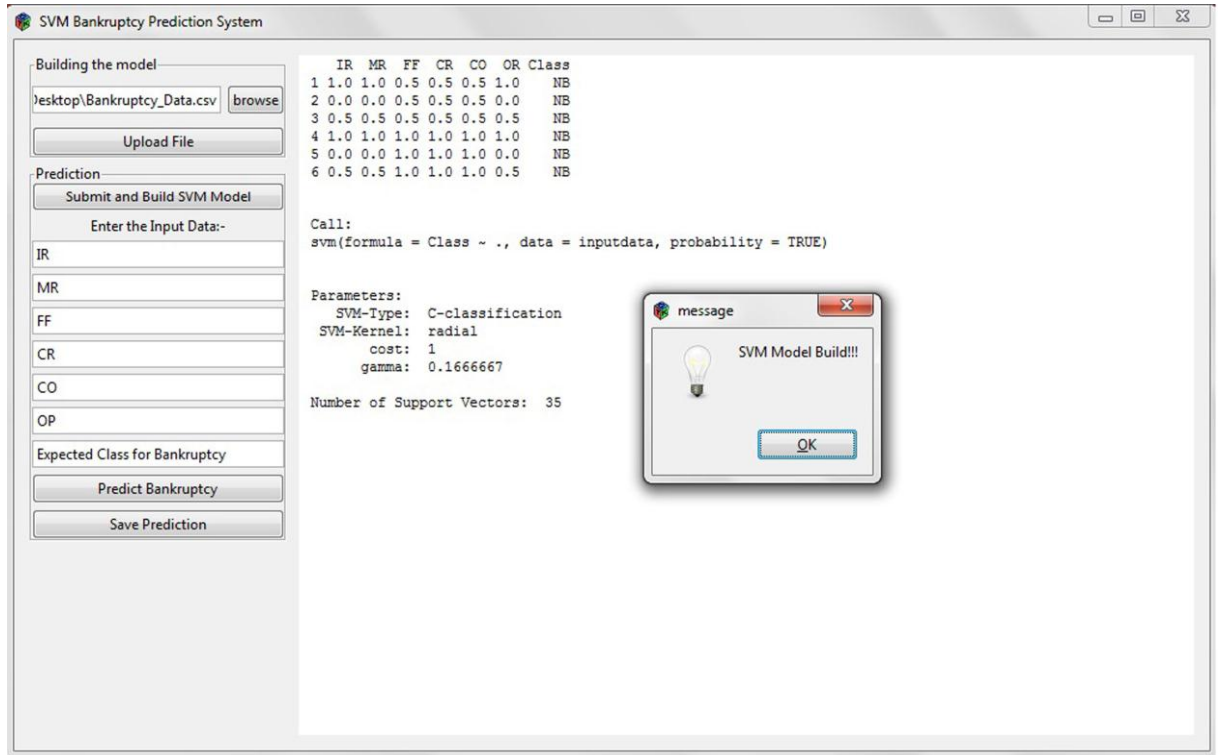

After the model is developed, users can enter their data in the text input boxes for predicting bankruptcy. Each of the six input parameters have values as 1, 0.5 or 0 (positive, average and negative) respectively. Based on SVM model built for the initial dataset, the predictive system estimates the probability of bankruptcy as either B (Bankruptcy) or NB (Non Bankruptcy). The predicted outcome is saved as a .csv file in the local directory of user's system to view the results. The output from the predictive system is shown in Figure 4, 5 and 6 respectively.

**5.5. Developing the prediction system as a package**

The predictive system for detecting bankruptcy was encoded as a package in RStudio. The package was developed using interdependent libraries 'devtools' and 'roxygen2'. The package can be downloaded by users in their local machines followed by installing and running the package in RStudio. Once the package is installed users can run the predictive system for detection of bankruptcy using their input data.

# 6. Conclusion

The results suggest that machine learning techniques can be implemented for prediction of bankruptcy. To serve the financial organizations for identifying risk oriented customers a prediction system was implemented. The predictive system helps to predict bankruptcy for a customer dataset based on the SVM model.